\documentclass[10pt,twocolumn,letterpaper]{article}

\usepackage[pagenumbers]{cvpr} 
\usepackage{booktabs, multirow}
\usepackage{caption}
\usepackage{subcaption}
\usepackage{amssymb}
\usepackage{pifont}
\usepackage{array,graphicx}
\usepackage[utf8]{inputenc}
\usepackage{url}
\usepackage{algorithm}
\usepackage{algpseudocode}
\usepackage{mathtools}
\usepackage{comment}

\newif\ifdraft
\drafttrue
\ifdraft

\newcommand{\smc}[1]{{\color{cyan}[\textbf{Sizhuo} #1]}}
\newcommand{\sm}[1]{{\color{black}#1}}

\else
\newcommand{\smc}[1]{}
\newcommand{\sm}[1]{{\color{black}#1}}
\fi

\definecolor{cvprblue}{rgb}{0.21,0.49,0.74}
\usepackage[pagebackref,breaklinks,colorlinks,allcolors=cvprblue]{hyperref}

\def\paperID{} 
\def\confName{CVPR}
\def\confYear{2026}

\title{gQIR: Generative Quanta Image Reconstruction}

\author{Aryan Garg \\
University of Wisconsin-Madison\\
{\tt\small agarg54@wisc.edu}
\and
Sizhuo Ma\\
Snap Inc.\\
{\tt\small sma@snapchat.com}
\and
Mohit Gupta\\
University of Wisconsin-Madison\\ 
{\tt\small mgupta37@wisc.edu}
}

\begin{document}
\maketitle

\vspace{-0.3in}
\begin{abstract}
Capturing high-quality images from only a few detected photons is a fundamental challenge in computational imaging. Single-photon avalanche diode (SPAD) sensors promise high-quality imaging in regimes where conventional cameras fail, but raw \emph{quanta frames} contain only sparse, noisy, binary photon detections. Recovering a coherent image from a burst of such frames requires handling alignment, denoising, and demosaicing (for color) under noise statistics far outside those assumed by standard restoration pipelines or modern generative models. 
We present an approach that adapts large text-to-image latent diffusion models to the photon-limited domain of quanta burst imaging. Our method leverages the structural and semantic priors of internet-scale diffusion models while introducing mechanisms to handle Bernoulli photon statistics. By integrating latent-space restoration with burst-level spatio-temporal reasoning, our approach produces reconstructions that are both photometrically faithful and perceptually pleasing, even under high-speed motion. We evaluate the method on synthetic benchmarks and new real-world datasets, including the first color SPAD burst dataset and a challenging \textit{Deforming (XD)} video benchmark. Across all settings, the approach substantially improves perceptual quality over classical and modern learning-based baselines, demonstrating the promise of adapting large generative priors to extreme photon-limited sensing. 
Code at \href{https://github.com/Aryan-Garg/gQIR}{https://github.com/Aryan-Garg/gQIR}.
\end{abstract}

\vspace{-5mm}

\begin{figure*}[t]
\begin{center}
    \centering
    \captionsetup{type=figure}
    \includegraphics[width=.96\textwidth]{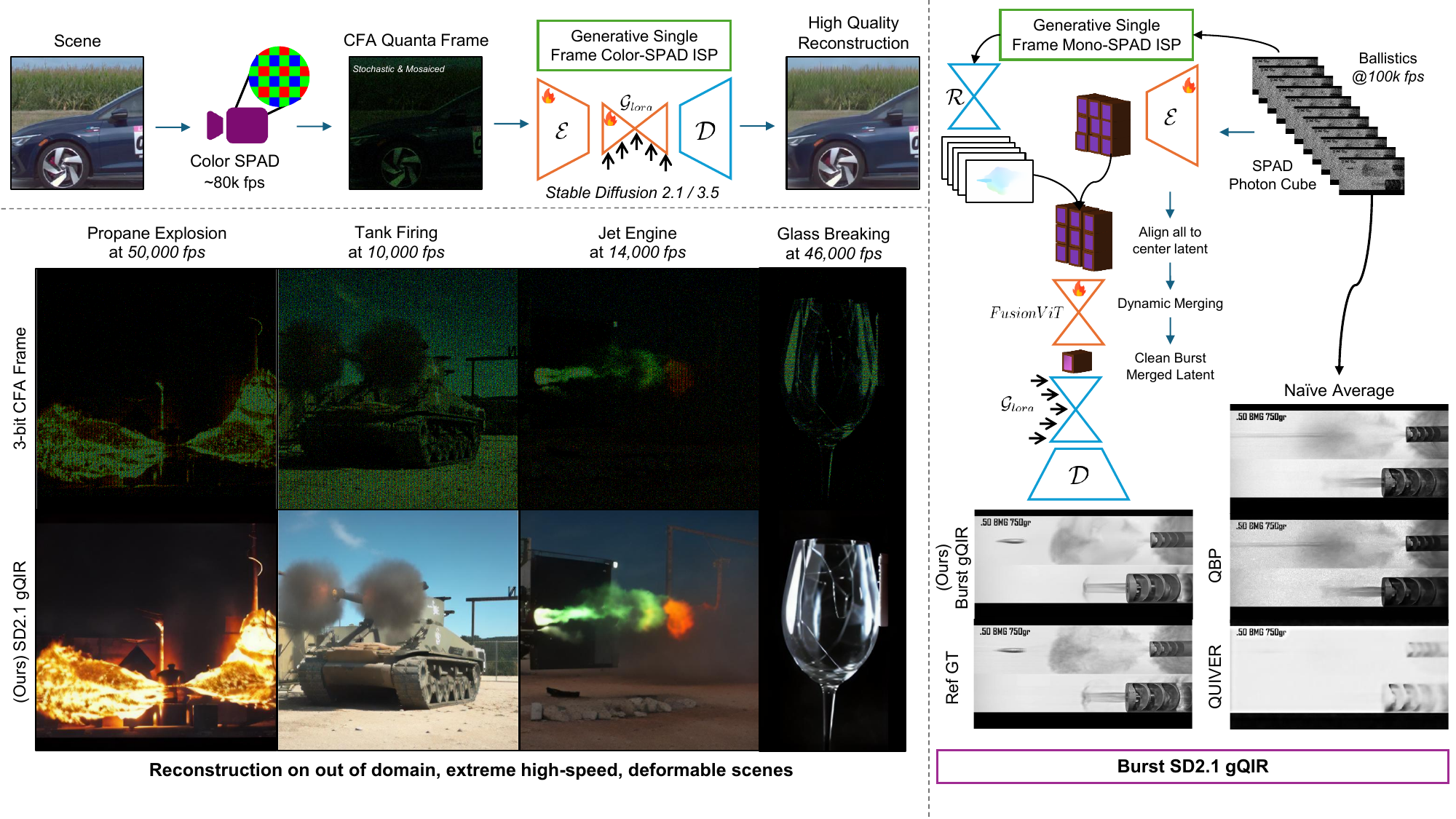}
    \vspace{-0.15in}\captionof{figure}{\textbf{gQIR: Photorealistic single image and burst reconstruction from ultra–high-speed color SPADs.} 
    Our pipeline reconstructs high-quality RGB images from 3-bit color-SPAD CFA nano-bursts (left) and merges SPAD photon cubes into temporally consistent bursts (right). 
    From photon-starved inputs captured at 10k–50k fps in extreme, out-of-domain scenes, gQIR recovers sharp textures, accurate color, and coherent structure by leveraging a generative prior. 
    For burst sequences up to 100k fps, FusionViT aligns and dynamically merges quanta latents, outperforming traditional and learning-based methods in both fidelity and perceptualness under motion.
}
    \label{fig:teaser}
    \vspace{-0.2in}
\end{center}%
\end{figure*}

\section{Introduction}\label{sec:intro}

Capturing a clear image from just a few detected photons is a long-standing challenge in computational imaging. Single-photon avalanche diode (SPAD) sensors~\cite{niclass_spad, spad_32Mpx} hold the promise of high-fidelity imaging in extreme low-light and high-speed regimes, where conventional sensors fail. However, each photon detection is a discrete stochastic event; as a result, individual \emph{quanta frames} are dominated by shot noise and quantization artifacts, often containing only sparse scene information~\cite{quanta_burst_photo, Fossum_single-bit_and_multi-bit}. Recovering a high-quality photograph from such frames requires combining information across a temporal burst of quanta frames.

SPAD arrays can operate in a Bernoulli mode, where each pixel records a binary value: 1 if one or more photons are detected during an exposure, and 0 otherwise. A short sequence of such binary frames—termed a \emph{nano-burst}—can be aggregated into a higher bit-depth representation (e.g., seven binary frames combined into a 3-bit frame). 
This burst-mode strategy enables photon-limited imaging at high frame rates and provides an opportunity to recover high-quality images from sparse photon events. However, aggregating these frames into a coherent image is challenging: small inter-frame motions cause misalignment, while the paucity of photons renders conventional motion estimation unreliable. This interplay between motion estimation, alignment, and photon-limited denoising defines the core challenge of \emph{quanta burst reconstruction}.

Early approaches to this problem relied on classical vision techniques~\cite{quanta_burst_photo, seeing_photons_in_color}, which explicitly estimate motion between frames. 
More recently, learning-based methods have been introduced~\cite{qudi, quiver}, which treat alignment and fusion as learnable modules. 
Despite this progress, photon-starved scenes with extreme deformation or ultra–high-speed motion remain challenging, as illustrated in \cref{fig:teaser}. Leveraging the representational power of large-scale generative models for quanta imaging remains an open direction. 
In particular, learning-based methods do not yet exploit the structural knowledge embedded in large text-to-image (T2I) diffusion models~\cite{stablediffusion_sd_original,sdxl,sdxl_lightning,sd3_sd35,chen2024pixartalpha,imagen,DALLE2,parti,CogView2}.
Generative restoration models~\cite{supir,instantIR,hypir,OSEDiff,resshift,diffBIR,star} leveraging such T2I priors have shown strong performance on conventional camera images. 
However, these models break down in the photon-limited regime for quanta cameras, where noise is non-Gaussian and photon counts far below those in conventional photography. \sm{Naive fine-tuning of these models leads to shortcut learning and does not produce meaningful outputs.} Furthermore, most prior quanta reconstruction methods consider monochrome sensors. 
In contrast, photon-counting color sensors~\cite{seeing_photons_in_color} introduce additional challenges due to sparse photon events in each color channel. 
\sm{Together, these factors make photon-limited burst reconstruction a demanding testbed for adapting large generative models to discrete, sparse quanta measurements.} 
As shown in \cref{fig:teaser}, the benefits of doing so emerge most clearly under extreme deformation or ultra–high-speed conditions.

To address these challenges, we propose a modular three-stage framework that adapts latent diffusion models for quanta burst reconstruction. The first stage jointly denoises and demosaics single or nano-burst quanta frames \sm{by finetuning the variational autoencoder (VAE) for latent space alignment} while mitigating catastrophic forgetting~\cite{catastrophic_forgetting}.
The second stage enhances perceptual fidelity through adversarial finetuning of the Low-rank adaptated (LoRA~\cite{hu2022lora}) latent U-Net. 
Finally, the third stage extends the framework to full bursts by generalizing the classical align-and-merge operation to latent space. 
A lightweight spatio-temporal transformer~\cite{miniViT} refines the center-frame latent using context from surrounding frames, thereby mitigating temporal artifacts such as flicker and content drift.

Our main contributions are as follows:
\begin{itemize}
    \item A modular approach that adapts large-scale T2I generative priors (e.g., Stable Diffusion~\cite{stablediffusion_sd_original}) to the extreme regime of quanta burst reconstruction.
    \item A learning-based method to jointly denoise, demosaic, and align bursts from color single-photon sensors, and a latent-space spatio-temporal transformer that enhances temporal consistency and mitigates content drift.
    \item The first real-world color SPAD burst dataset and a new \textit{eXtreme motion + Deforming (XD)} video dataset.
\end{itemize}

\smallskip\noindent 
This work takes a first step toward adapting large-scale generative priors to photon-limited sensing with quanta cameras, enabling high-quality color and monochrome reconstructions under ultra high-speed motion.

\begin{figure*}[t]
    \centering
    \includegraphics[width=0.95\linewidth]{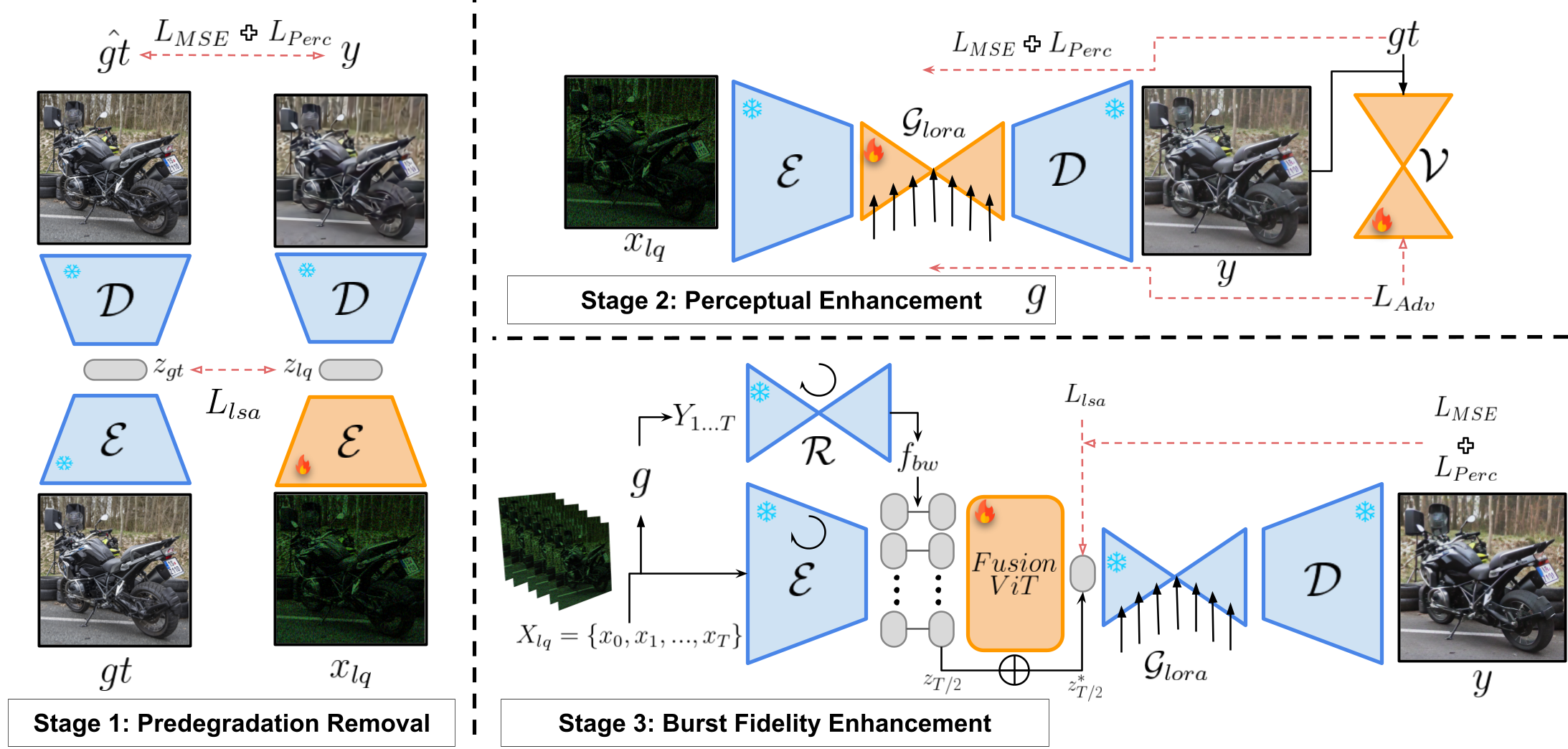}
    \caption{\textbf{Overview of gQIR.} Three-stage framework for quanta burst reconstruction: (S1) a quanta-aligned VAE for joint denoising and demosaicing of SPAD nano-bursts, (S2) an adversarially finetuned LoRA~\cite{hu2022lora} latent U-Net initialized with stable diffusion~\cite{stablediffusion_sd_original} weights for perceptual enhancement, and (S3) a latent burst FusionViT for motion-aware spatio-temporal fusion of burst of nano-burst inputs.}
    \label{fig:arch}
    \vspace{-0.15in}
\end{figure*}



\begin{figure*}[htpb]
    \centering
    \includegraphics[width=0.99\linewidth]{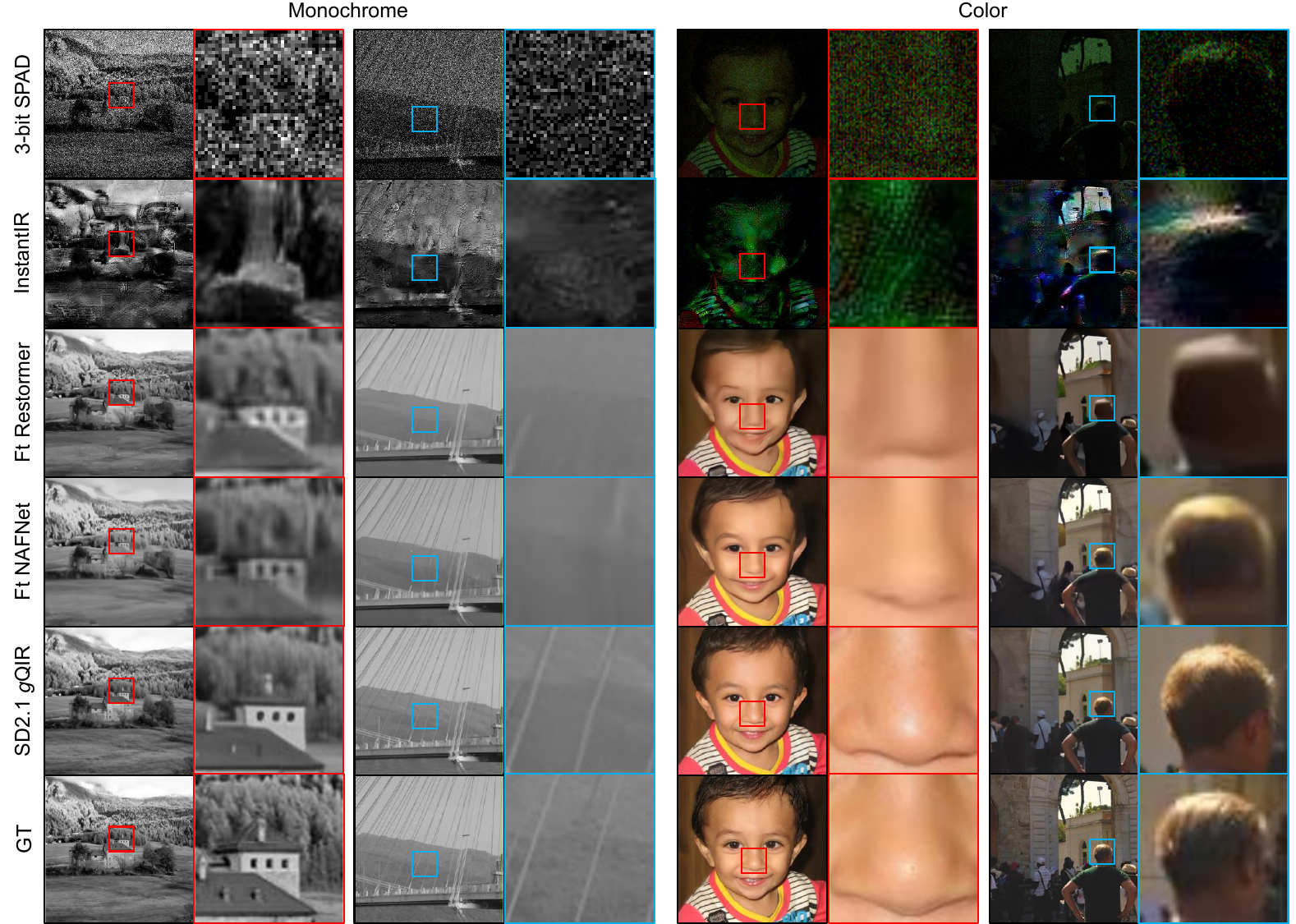}
    \caption{\textbf{Qualitative comparison – single 3-bit frame reconstructions.} Conventional finetuned baselines over-smooth high-frequency structures, especially in distant depth planes and textured regions, whereas gQIR preserves sharper details and more faithful facial features, benefitting from the inclusion of FFHQ faces~\cite{StyleGAN_ffhq} in the training set.}
    \label{fig:comp_single_fram}
    \vspace{-0.15in}
\end{figure*}
\section{Related Work}
\label{sec:related}

\noindent \textbf{Denoising for Conventional Cameras.}
State-of-the-art denoising networks~\cite{nafnet,Zamir2021Restormer} set strong baselines for conventional image restoration. 
NAFNet~\cite{nafnet} adopts a minimalist residual design without nonlinear activations, prioritizing spatial fidelity and efficiency, while Restormer~\cite{Zamir2021Restormer} employs transformer-based long-range modeling to capture complex noise characteristics. 
Although effective on Poisson–Gaussian noise typical of standard sensors, these models are not directly applicable to SPAD imagery, where photon shot noise, binary quantization, and extreme sparsity fundamentally alter the noise distribution. 
To provide representative baselines, we adapt and finetune both architectures for SPAD denoising and demosaicing.

\noindent \textbf{Quanta Burst Reconstruction.}
Quanta Burst Photography (QBP)~\cite{quanta_burst_photo} introduced a burst-denoising pipeline combining block-matching temporal alignment with frame merging and Wiener filtering. 
A subsequent work~\cite{seeing_photons_in_color} extended this framework to color, analyzing filter design for SPAD sensors and proposing a color-QBP variant to reconstruct RGB images from mosaiced bursts. 
Later reconstruction methods incorporated learned components while retaining the align-and-merge philosophy. 
QUIVER~\cite{quiver} uses an 11-frame window of 3-bit bursts, applying light pre-denoising to stabilize optical-flow estimation (SpyNet~\cite{spynet}) before recurrent fusion, while QuDI~\cite{qudi} replaces QUIVER’s recurrent denoiser with a time-conditioned U-Net, unrolling into a DDPM-like~\cite{ddpm} formulation. 
Work on multi-bit quanta and QIS reconstruction~\cite{image_reconstruction_qis_dnn,dynamic_low_light_qis} and on binary-to-multi-bit mappings~\cite{bit2bit} also use learned models for photon-efficient sensing. 
In parallel, efficiency-oriented approaches~\cite{Varun_generalizedEvents,Tianyi_StreamingPaper} reduce bandwidth by compressing quanta sequences before reconstruction, offering complementary benefits but addressing a goal orthogonal to reconstruction fidelity. 
Taken together, existing methods remain task-specific and operate without large-scale pretrained generative priors, a gap our approach addresses by adapting latent diffusion models to the quanta burst setting.

\noindent \textbf{Generative Image Restoration.}
Diffusion-based generative models~\cite{ddpm,ddim} introduced iterative denoising to learn natural image distributions, and large text-to-image (T2I) diffusion models~\cite{chen2024pixartalpha,sd3_sd35,sdxl,stablediffusion_sd_original,sdxl_lightning} have since become strong priors for a range of restoration tasks~\cite{wang2024sinsr,stableSR,yang2023pasd,wu2024seesr,OSEDiff,tsdSR,resshift}. 
Recent blind restoration methods~\cite{diffBIR,instantIR,supir,hypir} demonstrate that such priors can be adapted to diverse degradations while achieving high perceptual quality. 
These approaches typically align the latent space of a pretrained diffusion model to a target degradation via lightweight finetuning or adapters. 
Our VAE alignment stage (\cref{subsec:qaVAE}) follows this strategy but extends it to quanta sensors, where observations differ substantially from the continuous domains for which T2I priors are trained. 
To reduce the iterative sampling cost of diffusion models, we propose an adversarial finetuning stage (\cref{sec:stage2}), which, similar to~\cite{hypir}, produces a one-step generator while retaining the benefits of pretrained T2I priors.

\section{Methodology}
We describe our framework shown in~\cref{fig:arch} subsequently.



\subsection{Image Formation Model}
Photon imaging fundamentally differs from conventional cameras: each pixel registers discrete photon arrivals rather than analog intensities, \sm{with negligible read noise in SPADs.}
A physically consistent probabilistic rendering pipeline is used to synthesize a quanta observation from a clean sRGB ground truth image $x_{gt} \in [0,1]^{H\times W \times3}$. 
First, $x_{gt}$ is mapped to linear radiance space via 
$x_{lin}=x_{gt}^{\gamma}$, \sm{using a fixed $\gamma=2.2$}, so that pixel intensities scale proportionally with scene irradiance.
Given this linear image, a SPAD records whether at least one photon arrives during the exposure. 
Assuming Poisson arrivals with rate $\lambda$, the SPAD output $x_{spad}$  follows a Bernoulli distribution~\cite{quanta_burst_photo}:
\begin{equation}
    x_{spad} = Bern(1-e^{- \lambda})
    = Bern(1-e^{(- \alpha \cdot x_{lin})}),
    \label{eq:simulator}
\end{equation}
where 
$\lambda = \alpha \cdot x_{lin}$, with  $\alpha$ controlling the expected photon-per-pixel (PPP) level. 
The expected PPP is  $\mathbb{E}[\lambda]= \alpha \mathbb{E}[x_{lin}]$. 
In our setup, $\alpha=1.0$ or an average PPP of 3.5 matches the illumination levels in~\cite{quiver, qudi}.

To simulate a color SPAD, we apply a randomly sampled Bayer pattern $\pi \in \{\text{RGGB, GRBG, BGGR, GBRG}\}$, yielding a mosaiced binary frame $x_{lq}=M_{\pi}(x_{spad})$.
An $N$-frame or $log_2(N+1)$-bit mosaiced observation is:
\begin{equation}
    x_{lq} = \frac{1}{N} \sum_{i=1}^N M_{\pi} [Bern(1-e^{- \alpha\cdot x_{lin}})]
    \label{eq:sim_bayer_burst}
\end{equation}


\subsection{Stage 1: Quanta Aligned VAE}
\label{subsec:qaVAE}

\begin{figure}[t]
    \vspace{-0.1in}
    \centering
    \includegraphics[width=0.999\linewidth]{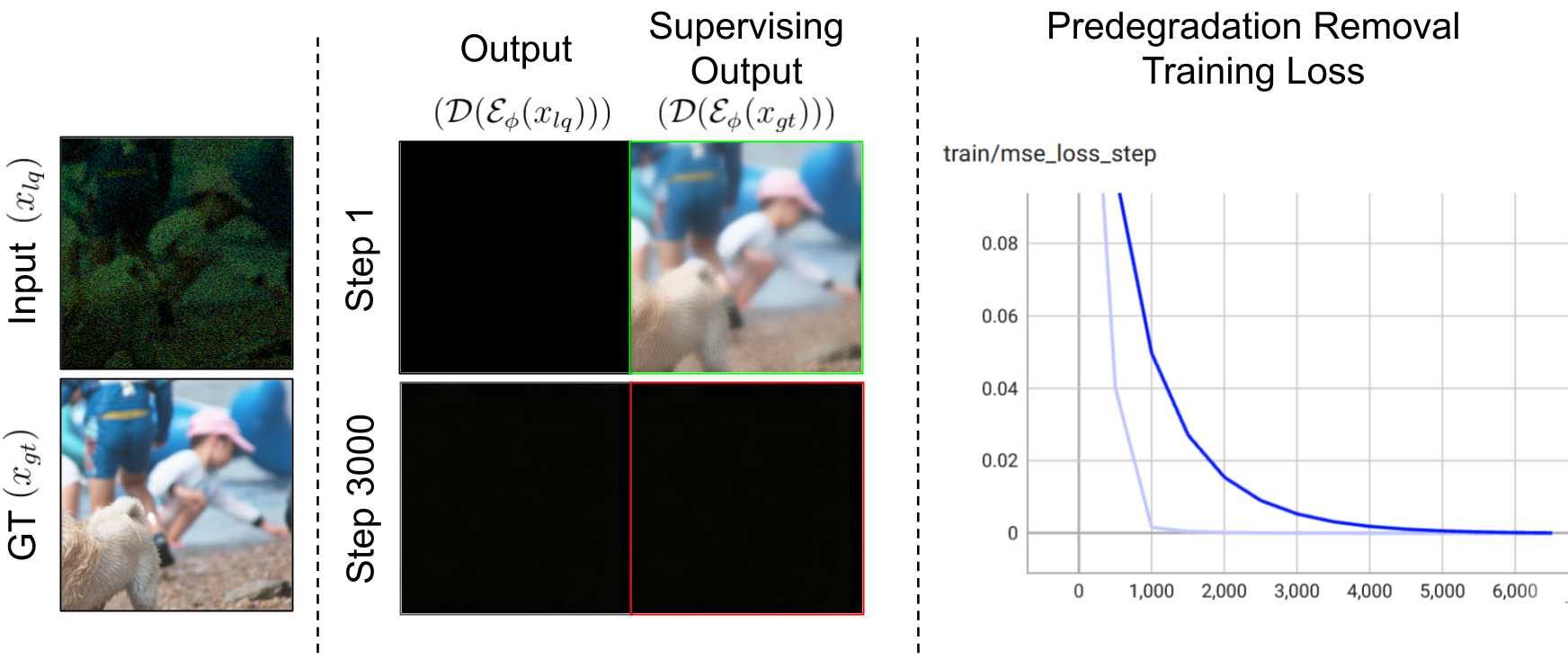}
    \caption{\textbf{Encoder collapse under predegradation removal loss~\cite{supir, hypir}.} The encoder $\mathcal{E}_{\phi^*}$ learns a perceptually meaningless shortcut thus producing constant outputs.
    Since the trainable encoder controls both, the supervision and prediction terms, the training curve quickly converges to a degenerate optimum without our proposed modifications.}
    \label{fig:predegradation_loss_ablation}
    \vspace{-0.15in}
\end{figure}

\noindent Generative restoration methods~\cite{supir, hypir, diffBIR} make their VAE's encoders degradation-aware by optimizing the objective: $\mathcal{L_{\mathcal{E}_{\phi^*}}} = \| \mathcal{D}(\mathcal{E}_{\phi^*}(x_{LQ})) - \mathcal{D}(\mathcal{E}_{\phi^*}(x_{GT}))\|_2^2$\sm{, where $\mathcal{D}$ denotes the frozen decoder, $\mathcal{E}_{\phi^*}$ denotes the finetuned encoder, and $x_{LQ}$, $x_{GT}$ denote the low-quality and ground truth images respectively}. 
This step is commonly known as degradation pre-removal~\cite{diffBIR, supir} which partially addresses restoration.
However, naively applying this step leads to catastrophic latent-space forgetting in our case due to the extreme photon-shot noise in SPADs. 
The encoder eventually finds a smoothed shortcut solution to generate the same image, regardless of the input, as shown in~\cref{fig:predegradation_loss_ablation}. 
To address this, we introduce \sm{two key modifications: \emph{deterministic mean encoding} and \emph{latent space alignment loss.}}

\noindent \sm{\textbf{Deterministic Mean Encoding.}} Instead of stochastic sampling from the posterior 
$q_{\phi}(z | x_{lq}) = \mathcal{N}(\mu_{\phi}(x_{lq}), \sigma_{\phi}^2(x_{lq}) )$ we use the deterministic mean: $\mathbb{E}_{q_{\phi} (z | x_{lq}) } =\mu_{\phi}(x_{lq})$ obtained from the frozen, pre-trained encoder $\mathcal{E}_{\phi}$. 
This deterministic formulation avoids stochastic variance amplification, which is particularly important since $x_{lq}$ is severely corrupted by photon-shot noise and already exhibits heavy-tailed statistics. 
Our objective is to maximize fidelity by preserving the latent structure of the underlying clean scene. 
We achieve this by adding our new latent loss to modified reconstruction losses described as follows.

\noindent \textbf{Latent Space Alignment (LSA) Loss.}
Since the decoder remains fixed throughout training to preserve its internet-scale training learned manifold, the encoder ($\mathcal{E_{\phi^*}}$) must perform simultaneous denoising and demosaicing to produce a clean latent representation from $x_{lq}$.
We enforce latent consistency between the low-quality and ground-truth embeddings using the following alignment loss:
\begin{equation}
\mathcal{L}_{lsa} = \| \mu_{\phi^*}(x_{lq}) - \mu_{\phi}(x_{gt}) \|_2^2.
\label{eq:lsa_loss}
\end{equation}
It is worth noting that the second term, unlike~\cite{supir, hypir} utilizes a frozen copy of the pre-trained Encoder $E_{\phi}$. 
This safeguards against the predegradation removal encoder's collapse as shown in~\cref{fig:predegradation_loss_ablation}.

\noindent \textbf{Pixel Space Losses.}
We also use MSE and LPIPS loss~\cite{lpips}:
\begin{equation}
    \mathcal{L}_{MSE} =  \| \mathcal{D} ( \mu_{\phi^*}(x_{lq})) - \mathcal{D} (\mu_{\phi} (x_{gt})) \|_2^2,
\label{eq:reconstrucion_mse}
\end{equation}
\begin{equation}
    \mathcal{L}_{perc} =  \| \Phi(\mathcal{D} ( \mu_{\phi^*}(x_{lq}))) - \Phi(\mathcal{D} (\mu_{\phi} (x_{gt}))) \|_2^2,
\label{eq:reconstrucion_lpips}
\end{equation}
where $\Phi$ denotes a VGG-19~\cite{vgg} backbone. 

\noindent \textbf{Overall loss} is given by:
\begin{equation}
\begin{split}
     \mathcal{L}_{\mathcal{E}_{qvae}} = \lambda_{lsa} \mathcal{L}_{lsa} \, \, + \lambda_{MSE} \mathcal{L}_{MSE} \, \, + \lambda_{perc} \mathcal{L}_{perc},
\end{split}
\end{equation}
where $\lambda_{lsa}$, $\lambda_{MSE}$ and $\lambda_{perc}$ are scalar hyperparameters. 


\subsection{Stage 2: Perceptual Enhancement}
\label{sec:stage2}
\sm{Stage 1’s VAE alignment enables joint denoising and demosaicing, recovering structural, chromatic, and low-frequency details. 
In Stage 2, we finetune the pretrained diffusion backbone to refine the reconstruction, enhancing high-frequency details and improving  perceptual quality.

Due to the extremely high data capture rate of SPAD sensors, reconstruction algorithms are often faced with huge amount of data processing, motivating the design of single-step algorithms. 
Adversarial training has been established as an effective way of distilling the diffusion prior to a single-step model~\cite{advDist001_kang2024diffusion2gan, advDist002_yin2024onestep, advDist003_yin2024improved, advDist004_ADD, sdxl_lightning}. 
Specifically, \cite{hypir} demonstrated a theoretical guarantee for stable GAN training by initializing the LoRA-initialized denoising network $\mathcal{G}_{lora}$ with the prior's diffusion weights $\mathcal{U_{\phi}}$, which ensures small initial gradients for a stable start of GAN training.
We design a multilevel ConvNext-Large~\cite{convNext_Large} backbone discriminator $\mathcal{V_{\theta}}$ modified from~\cite{hypir} to adversarially train $\mathcal{G}_{lora}$ using the standard min-max GAN objective~\cite{gan}:
\begin{equation}
    \min_{\phi} \max_{\theta} \mathbb{E}_{x \sim p_{X_{gt}}}[\log \mathcal{V_{\theta}}(x)] + \mathbb{E}_{x \sim p_{X_{lq}}}[\log(1-\mathcal{V}_{\theta}(\mathcal{G}(x)))] .
\label{eq:adv}
\end{equation}
The generator is additionally updated~\cite{gan_reconstruction} by the pixel space reconstruction and the perceptual loss. Overall:
\begin{equation}
    \begin{split}
    \mathcal{L}_{{G_{lora}}} = \mathcal{L}_{adv} +  L_{perc} +  \| \mathcal{D} ( G_{lora}( \mu_{\phi^*}(x_{lq}))) - x_{gt}\|_2^2 .
    \end{split}
    \label{eq:mse_perceptual_generator}
\end{equation}
}

\begin{figure}[t]
    \centering
    \includegraphics[width=0.99\linewidth]{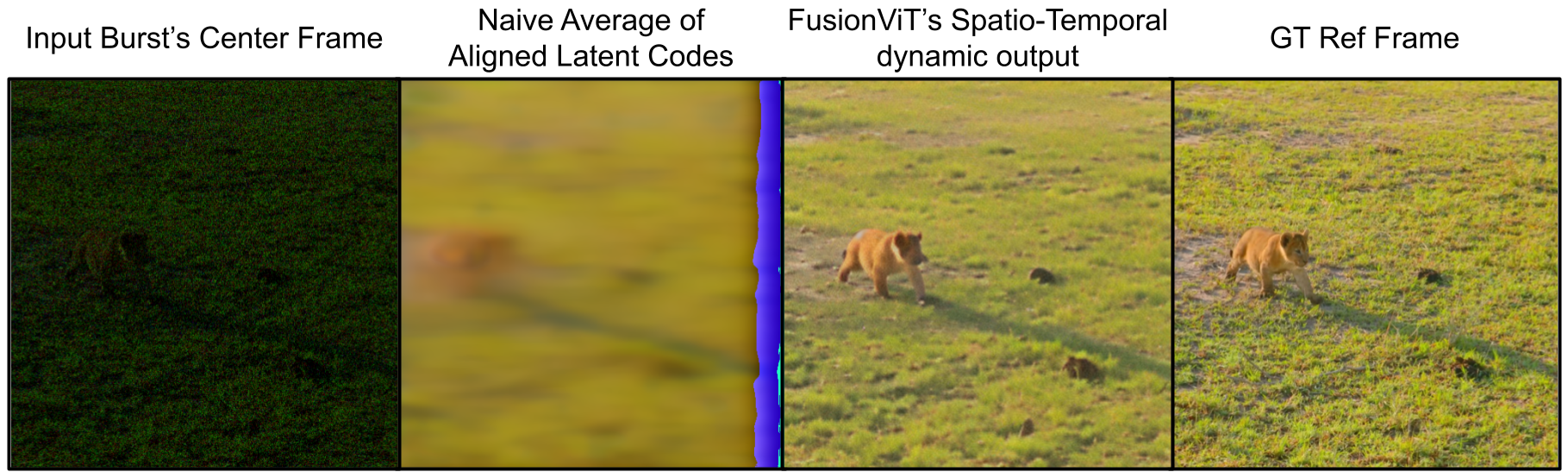}
    \caption{\textbf{Dynamic spatio-temporal latent burst merging.} Naive averaging of flow-aligned burst latents yields blur under scene motion. FusionViT instead adaptively weights latents by motion and proximity to the reference, producing a sharper output.}
    \label{fig:naive_avg_latents_vs_fusion_vit}
\vspace{-0.15in}
\end{figure}

\begin{figure*}[ht]
    \centering
    \includegraphics[width=0.99\linewidth]{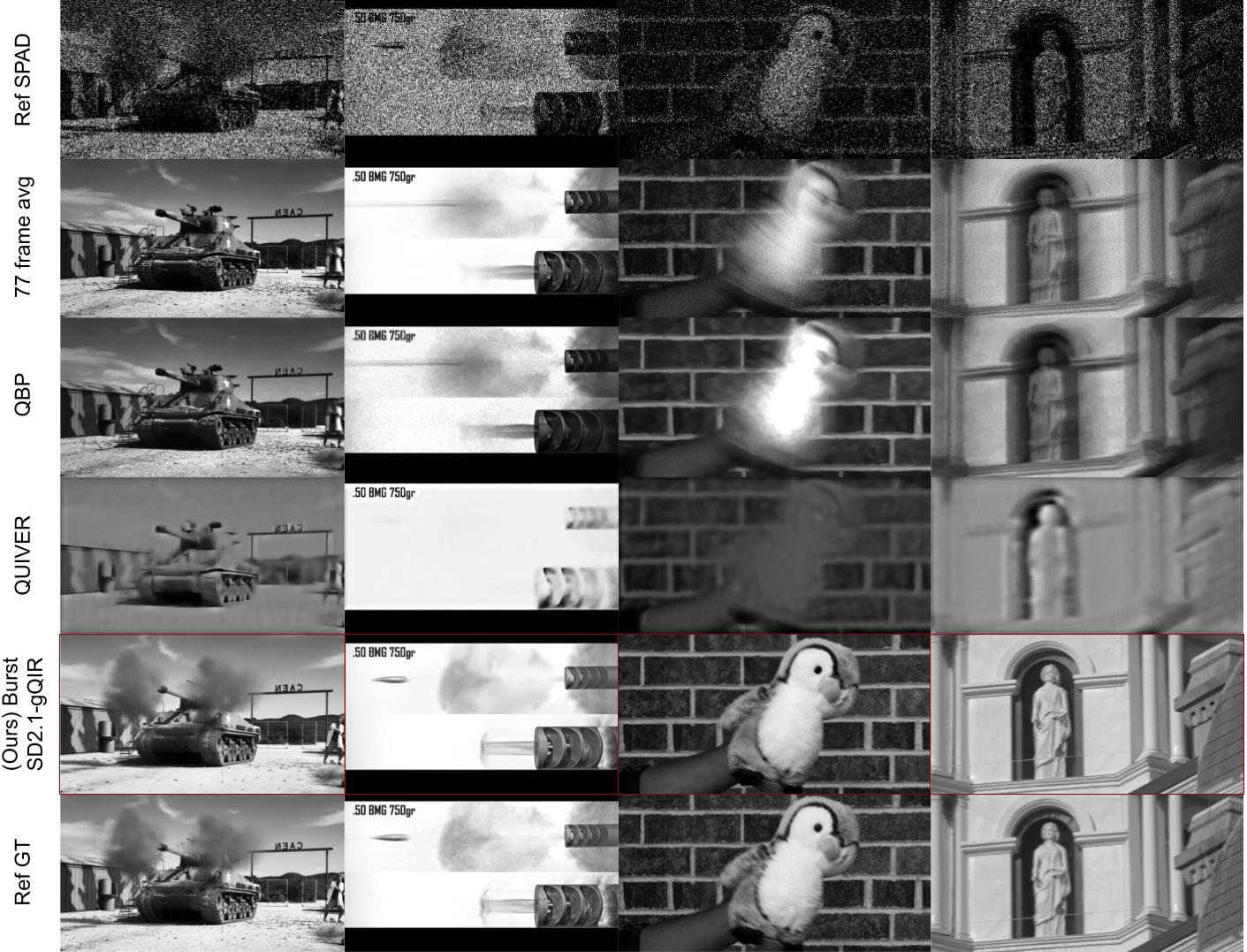}
    \caption{\textbf{Qualitative comparison – burst reconstruction.} We simulate 1:1 GT–SPAD bursts by averaging 77 binary frames per input, preserving the original scene frame rate. 
    QBP yields blurred reconstructions under fast motion due to small burst input while QUIVER breaks down due to motion-blurred nano-bursts, from realistic sampling. 
Our burst pipeline consistently recovers sharper structure and higher fidelity across extreme motion regimes, from 1000 to 100k fps.}

    \label{fig:burst_comp}
    \vspace{-0.15in}
\end{figure*}

\vspace{-4mm}
\subsection{Stage 3: Latent Burst Imaging}
Next, we extend our model with a burst window~\cite{quanta_burst_photo, quiver, qudi} to exploit temporal information in a quanta burst sequence.
We generalize the align-and-merge philosophy of QBP~\cite{quanta_burst_photo} to the VAE's latent space.   We first compute optical flow to align all burst latent maps to the center latent map $z_{c}$.
However, a pre-trained optical flow estimator $\mathcal{R}$ cannot accurately warp the latent map despite the presence of rich semantic information.
Moreover, applying pre-trained models directly to the low-quality sequence $X = (x_{lq}^0, ... x_{lq}^i, x_{lq}^{i+1} ...)$ fails due to a significant domain gap  (See supplementary). 
To bridge the gap, we first reconstruct all frames: $Y = \mathcal{D_{\phi}(}\mathcal{G}_{lora}(\mathcal{E_{\phi^*}}(X_{lq})))$ and then use  RAFT~\cite{raft}, pretrained on FlyingThings3D~\cite{SceneFlow_Datasets}, to estimate the flow $\mathcal{O} = \mathcal{R}(Y)$, similar in spirit to pre-denoising~\cite{quiver, qudi} or temporal aggregation~\cite{quanta_burst_photo}.



Once optical flow is estimated, all burst frames are warped to the reference and merged. 
Naively averaging these aligned latents produces significant blur due to motion (\cref{fig:naive_avg_latents_vs_fusion_vit}).
To overcome this, we introduce a pseudo-3D miniViT~\cite{miniViT} ($\mathcal{F}$) that applies sub-quadratic windowed attention across time and the spatial axis, enabling dynamic spatio-temporal burst fusion into a single high-fidelity latent code. Furthermore, the output of the FusionViT is modulated and residually added to the center latent $z_{T/2}$ as shown in~\cref{fig:arch}. 
The modulation is a learned scalar $\delta$, initialized at $0.05$, to adaptively add or subtract the fused details prior to feeding the latent code into the generative network $\mathcal{G}_{lora}$. 
We freeze all other networks and supervise FusionViT with the following overall loss, similar to Stage 1:
\begin{equation}
\begin{split}
    \mathcal{L}_{fusion} = \| \mathcal{F}(\mu_{\phi^*}(X_{lq})) - \mu_{\phi}(x_{gt}) \|_2^2 + \\
    \|  \mathcal{D}(\mathcal{G}_{lora}(\mathcal{F}(\mu_{\phi^*}(X_{lq})))) - x_{gt}\|^2_2 + \mathcal{L}_{perc}
\end{split}
\end{equation}


\subsection{Implementation Details}
\noindent \textbf{GT-SPAD Simulator}
We use $\alpha{=}1.0$ for all simulations (expected PPP: 3.5) with randomized Bayer pattern per iteration. 
Stages 1–2 use nano-bursts formed by averaging 7 independently sampled binary frames per GT while Stage 3 uses 1 binary frame per GT, yielding 11 3-bit nano-bursts from $(11{\times}7)\,\, 77$ binary frames.

\noindent \textbf{Datasets}
We train on 2.81 M images and 44,575 videos combined from diverse image~\cite{div2k, flickr2K, alis_landscapedHQ, StyleGAN_ffhq, laion-5B_laionHQ} and video datasets~\cite{reds_120fps_dataset, youHQ, visionsim, xvfi}.
Testing also adds UDM~\cite{udm10}, SPMC~\cite{spmc_dataset}, our eXtreme-Deformable (XD) dataset to the test-splits of the aforementioned.
See supplementary.

\noindent \textbf{Hyperparameters and Training.}
Stage 1 (SPAD–GT alignment VAE) is trained for 600k steps on 8$\times$A100 with LR $10^{-5}$, batch size 8, and scalars: $\lambda_{lsa}{=}0.1$, $\lambda_{MSE}{=}10^3$, $\lambda_{perc}{=}2$.  
Stage 2 runs for 100k iterations on a single RTX 4090 at 256$\times$256 with losses $\lambda_{adv}{=}0.5$, $\lambda_{MSE}{=}500$, $\lambda_{perc}{=}5$.  
Stage 3 (FusionViT) trains for 20k steps using RAFT~\cite{raft} pretrained on FlyingThings3D~\cite{SceneFlow_Datasets}, with $\lambda_{lsa}{=}1$, $\lambda_{MSE}{=}1000$, $\lambda_{perc}{=}7.5$.  
All stages use Adam~\cite{adam} optimizer with LR $\eta=10^{-5}$ and $\beta{=}(0.9,0.999)$. All implementations are in PyTorch~\cite{pytorch}.

\begin{table*}[htbp]
    \centering
    \footnotesize
    \setlength{\tabcolsep}{3pt} 
    \caption{\textbf{Fidelity and perceptual quality of 3-bit nano-burst input single RGB frame reconstruction.}
    Fine-tuned Restormer and NAFNet attain higher PSNR due to optimization for lower distortion~\cite{blau2018perception},
    leading to oversmoothing, while gQIR achieves higher perceptual quality, consistent with visual results in~\cref{fig:comp_single_fram}.}
    \vspace{0.3em}
    \begin{tabular}{l ccc ccc ccc ccc}
        \toprule
        & \multicolumn{6}{c}{\textbf{Monochrome}} & \multicolumn{6}{c}{\textbf{Color}} \\
        \cmidrule(lr){2-7} \cmidrule(lr){8-13}
        \textbf{Method} &
        \multicolumn{3}{c}{\textbf{Full-Reference}} & \multicolumn{3}{c}{\textbf{Non-Reference}} &
        \multicolumn{3}{c}{\textbf{Full-Reference}} & \multicolumn{3}{c}{\textbf{Non-Reference}} \\
        \cmidrule(lr){2-4} \cmidrule(lr){5-7} \cmidrule(lr){8-10} \cmidrule(lr){11-13}
        & PSNR$\uparrow$ & SSIM$\uparrow$ & LPIPS$\downarrow$ &
          ManIQA$\uparrow$ & ClipIQA$\uparrow$ & MUSIQ$\uparrow$ &
          PSNR$\uparrow$ & SSIM$\uparrow$ & LPIPS$\downarrow$ &
          ManIQA$\uparrow$ & ClipIQA$\uparrow$ & MUSIQ$\uparrow$ \\
        \midrule
        InstantIR~\cite{instantIR} 
            & 10.787 & 0.178 & 0.651 & 0.187 & 0.346 & 36.651
            & 7.928  & 0.101 & 0.736 & 0.197 & 0.358 & 32.211 \\
        ft-Restormer~\cite{Zamir2021Restormer} 
            & \textbf{28.728} & 0.816 & 0.294 & 0.262 & 0.435 & 40.439
            & 26.433 & 0.739 & 0.388 & 0.235 & 0.395 & 36.026 \\
        ft-NAFNet~\cite{nafnet} 
            & 28.276 & 0.830 & \textbf{0.261} & 0.300 & 0.473 & 39.129
            & \textbf{26.881} & 0.757 & \textbf{0.338} & 0.251 & 0.431 & 36.732 \\
        (Ours) qVAE 
            & 28.180 & \textbf{0.863} & 0.327 & 0.299 & 0.487 & 44.895
            & 26.280 & \textbf{0.791} & 0.435 & 0.272 & 0.432 & 38.613 \\
        (Ours) gQIR 
            & 27.281 & 0.839 & 0.318 & \textbf{0.331} & \textbf{0.547} & \textbf{45.614}
            & 25.480 & 0.766 & 0.361 & \textbf{0.313} & \textbf{0.490} & \textbf{42.038} \\
        \bottomrule
    \end{tabular}
    \label{tab:quant_eval}
\end{table*}

\section{Experiments}
We first define the methods and experimental settings:

\noindent \textbf{Baselines.} 
We establish single-frame denoising baselines by finetuning two representative RGB methods, NAFNet~\cite{nafnet} and Restormer~\cite{Zamir2021Restormer} \sm{since no other baselines exist for single quanta
image reconstruction task, to the best of our knowledge.}
We do not finetune InstantIR~\cite{instantIR} as it relies on a pre-trained generative prior and is designed for test-time unknown degradation removal.
For the burst stage, we compare with quanta baselines: QBP~\cite{quanta_burst_photo} and QUIVER~\cite{quiver}, with a burst size of 3-bit 11 frames.

\noindent \textbf{Metrics.} We evaluate all methods using full-reference: PSNR, SSIM~\cite{ssim}, LPIPS~\cite{lpips} and non-reference metrics: ManIQA~\cite{manIQA}, ClipIQA~\cite{clipIQA} and MUSIQ~\cite{MUSIQ}. 
For burst comparisons, we focus on video temporal consistency reconstructed via sliding burst windows. 
$E_{warp}$~\cite{E_star} is used as the flow-warping metric ($E^* = 10^3E_{warp}$).


\noindent \textbf{Test Datasets.} We use two different sets curated from the aforementioned test-splits. 
The single image reconstruction test set consists of 334 images while the burst test set consists of 11 100-frame videos from XVFI-test, I2-2000fps and XD-Dataset. 
We also evaluate our burst method on the entire test split of I2-2000fps.

\subsection{Quantitative Evaluation}
\noindent \textbf{Single Image Comparisons.}
We provide quantitative evaluations of fidelity and perceptual scores on our single quanta image reconstruction task in~\cref{tab:quant_eval}. 
All methods are run at the finetuned baselines' $384^2$ resolution. 
\sm{It is well established that a trade-off exists between perceptual quality and distortion~\cite{blau2018perception}. While existing denoising methods optimize for lower distortion as evidenced by full-reference metrics, the input 3-bit frame contains extremely sparse information, and optimizing for distortion leads to oversmoothing. In contrast, by leveraging a strong generative prior, gQIR achieves superior perceptual quality, reflected in no-reference metrics and consistent with the qualitative comparisons in \cref{subsec:qualitative}.} 

\noindent \textbf{Burst Comparisons.}
We curate a subset of our entire test set, grouping sequences by frame rate to evaluate methods across a wide spectrum, unlike prior works with fixed fps~\cite{quiver, qudi}. Specifically, we select 11 sequences from XVFI~\cite{xvfi} (1k fps), I2-2000fps~\cite{quiver} (2k fps), and XD (2k–100k fps). 
\sm{\cref{tab:quant_eval_burst} shows that Burst-gQIR consistently outperforms the baselines, particularly on the challenging XD dataset, where QBP and QUIVER exhibit substantial performance drop.}

\sm{We further evaluate our method on the full I2-2000fps~\cite{quiver} test set as shown in \cref{tab:steal_qudi_evals}. Despite a minor domain gap between a PPP of 3.5 (ours) and 3.25~\cite{quiver}, our method surpasses the previous best (QuDI~\cite{qudi}) by +2.17 dB. }


\subsection{Qualitative Evaluation}
\label{subsec:qualitative}
\noindent \textbf{Single Image Comparisons.}
\sm{Monochrome and color SPAD reconstruction comparisons are shown in \cref{fig:comp_single_fram}. InstantIR fails to produce high-quality reconstructions due to the mismatch between Poisson–Gaussian and Bernoulli noise statistics. Fine-tuned Restormer and NAFNet yield reasonable but over-smoothed results, whereas the proposed gQIR restores fine details and enhances perceptual quality.}

\noindent \textbf{Burst Comparisons.}
\sm{\cref{fig:burst_comp} presents qualitative comparisons for burst reconstruction. While QUIVER~\cite{quiver} is trained and evaluated on motion-blur-free nano-bursts generated by inflating 11 GT frames into 77 binary sequences (sampling 7 binary frames per GT), we adopt a more realistic setting by sampling a single binary frame per GT, introducing motion blur in each nano-burst. Due to this domain gap, QUIVER fails to produce high-quality reconstructions. QBP, designed for bursts with hundreds of frames, yields blurry outputs under fast motion. In contrast, Burst-gQIR delivers sharp, high-fidelity reconstructions, effectively handling motion while preserving perceptual quality.}
\begin{table*}[ht]
    \centering
    \footnotesize
    \caption{\textbf{Burst reconstruction fidelity under extreme motion.}  Our method achieves superior scores due to cleaner flow procsesing and dynamic burst merging while keeping the traditional align-and-merge philosophy aided with a generative prior.}
    \begin{tabular}{l|ccc|ccc|ccc}
        \toprule
         \multicolumn{1}{c}{} & \multicolumn{3}{c}{QBP~\cite{quanta_burst_photo}} & \multicolumn{3}{c}{QUIVER~\cite{quiver}} & \multicolumn{3}{c}{\textbf{Burst-gQIR}} \\
        \cmidrule(lr){2-4} \cmidrule(lr){5-7} \cmidrule(lr){8-10}
        Test-set (GT fps) & PSNR$\uparrow$ & SSIM$\uparrow$ & LPIPS $\downarrow$ & PSNR$\uparrow$ & SSIM$\uparrow$ & LPIPS $\downarrow$ & PSNR$\uparrow$ & SSIM$\uparrow$ & LPIPS $\downarrow$ \\   
        \midrule
        
        XVFI~\cite{xvfi} (1000)    & 12.014 &  0.370  & 0.647  & 23.001 & 0.751 & 0.575 & \textbf{25.822} & \textbf{0.712} & \textbf{0.434}  \\ 

        I2-2000fps~\cite{quiver} (2000)  & 16.043 & 0.549 & 0.468 & 25.060 & 0.874 &  0.366 & \textbf{31.214} & \textbf{0.878} & \textbf{0.296} \\

        XD (2k - 100k) & 12.780 & 0.409 & 0.458 & 20.096 & 0.790 & 0.421  & \textbf{30.331} & \textbf{0.895} & \textbf{0.316} \\
    
        \midrule
            Cumulative:  & 13.380 & 0.448 &  0.496 & 22.429 & 0.814 &  0.429 & \textbf{29.832} & \textbf{0.856} & \textbf{0.330} \\
        \bottomrule
    \end{tabular}
    \label{tab:quant_eval_burst}
\end{table*}

\begin{table}[t]
    \footnotesize
    \centering
    \caption{\textbf{Burst Fidelity on I2-2k benchmark.}
    Despite the PPP mismatch, our method reaches superior fidelity on I2-2k~\cite{quiver}.}
    \begin{tabular}{lcc}
        \toprule
        Method & PSNR $\uparrow$ & SSIM $\uparrow$\\ 
        \midrule
        
        EMVD~\cite{emvd}                & 20.019	& 0.587  \\ 
        FloRNN~\cite{flornn}       & 21.034	& 0.679  \\ 
        QBP~\cite{quanta_burst_photo} 	 & 21.548	& 0.703  \\ 
        Transform Denoise~\cite{transform_denoise} & 21.317 & 0.718 \\ 
        QUIVER~\cite{quiver}      & 26.214	& 0.790  \\ 
        QuDi~\cite{qudi} 	             & 28.641   & 0.811 \\ 
        \midrule
        (Ours) Burst-gQIR	(3.25 PPP) & \textbf{30.811}  & \textbf{0.868} \\ 
        \bottomrule
    \end{tabular}
    \label{tab:steal_qudi_evals}
\end{table}





\begin{figure}[htbp]
    \centering
    \includegraphics[width=0.99\linewidth]{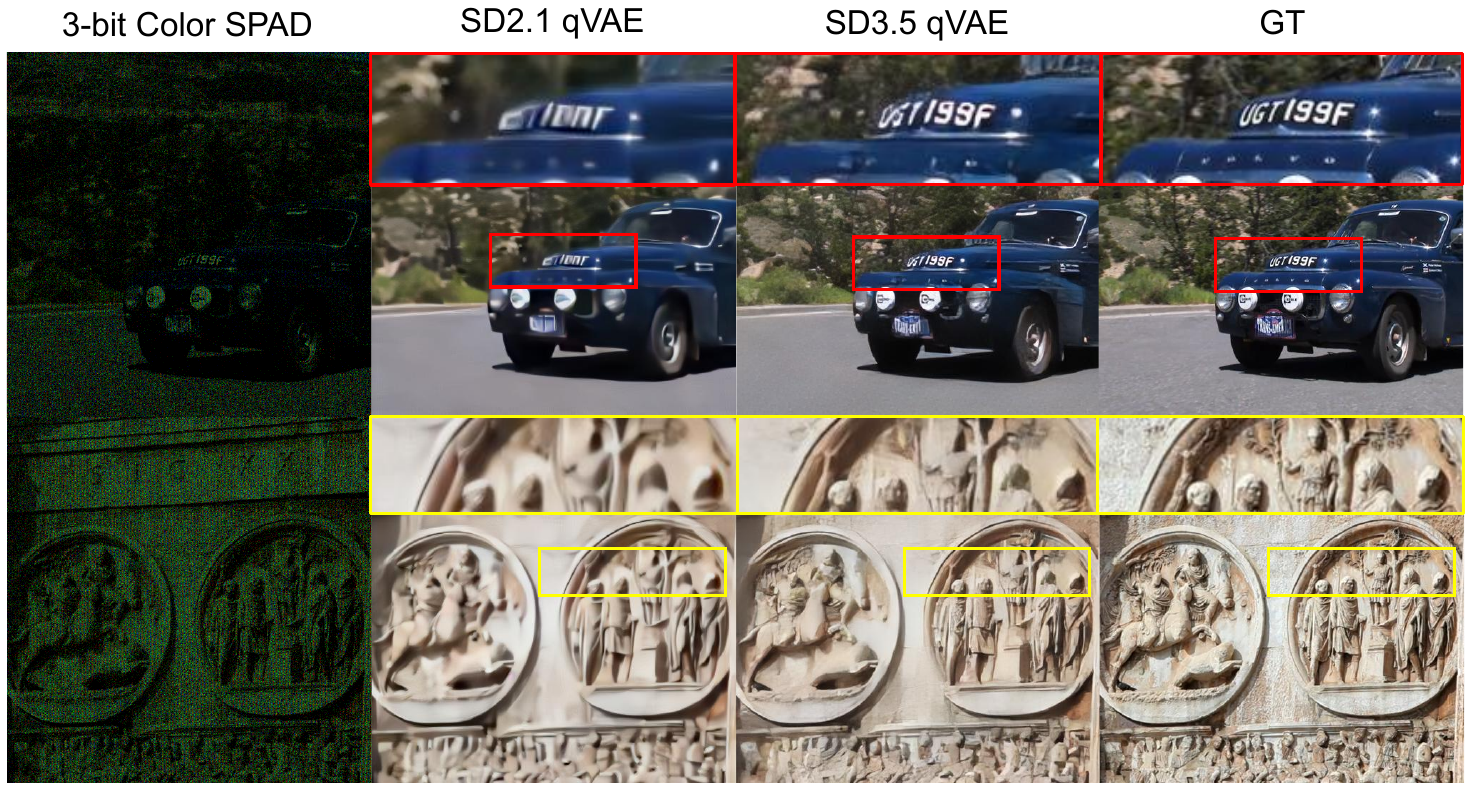}
    \caption{\textbf{SD3.5's VAE solves the text problem~\cite{textdiffuser, textdiffuser2}.} SD3.5~\cite{sd3_sd35} uses a $4\times$ larger latent space than SD2.1. This yields sharper high-frequency details and legible text drawing capabilities.}
    \label{fig:vae_ablation}
\end{figure}

\subsection{Ablation Studies}
\noindent \textbf{Stage 1 Design Choices.}
We ablate the introduced combination of Latent Space Loss (LSA) and deterministic sampling for quanta frames and observe that LSA provides a critical gradient to the encoder for convergence in~\cref{tab:stage1_losses_encoding}.
\begin{table}[tpb]
    \footnotesize
    \centering
    \caption{\textbf{Ablation - Stage 1 Design Choices and Losses.} 
    Our latent space alignment loss and deterministic sampling gives the highest fidelity in 1 epoch for joint denoising and demosaicing. 
    Both components are critical for meaningful convergence and avoiding catastrophic forgetting shown in \cref{fig:predegradation_loss_ablation}.}
    \begin{tabular}{lccc}
    \toprule
        Variants & PSNR $\uparrow$ & SSIM $\uparrow$ & ManIQA $\uparrow$  \\ 
    \midrule
         \textit{w/o} det. encoding (A) & 20.56 & 0.435 & 0.167 \\ 
         \textit{w/o} LSA loss (B) &  10.39 &  0.222 & 0.139\\ 
         \textit{w/o} (A) and (B)  & 10.30 &  0.218 & 0.136  \\ 
        \textbf{Ours} & \textbf{24.78} &  \textbf{0.665} & \textbf{0.194} \\ 
    \bottomrule
    \end{tabular}
    \label{tab:stage1_losses_encoding}
\end{table}

\noindent \textbf{VAE Scaling Solves the Text Problem~\cite{textdiffuser, textdiffuser2}.}
Increasing the VAE's latent dimensionality, markedly improves capacity thereby enabling text synthesis. 
We show a $4\times$ larger aligned SD3.5~\cite{sd3_sd35} qVAE in \cref{fig:vae_ablation} and the supplementary.

\noindent \textbf{Fidelity, Perceptualness and Video Stability.}
\sm{We compare all three stages of our method based on fidelity, perceptual quality, and video stability over the video test set used in~\cref{fig:burst_comp}.
Stage 2 enhances photorealism but incurs a higher degree of content drift. This is attributed to its greater emphasis on perceptualness during training. In contrast, Stage 3 is optimized for combining temporal information for higher fidelity. 
This naturally mitigates content drift, as demonstrated in~\cref{tab:s1vs2vs3_ablation}.
Qualitative video reconstruction comparisons are provided in the supplementary.}

\begin{table}[tpb]
    \footnotesize
    \centering
       \caption{\textbf{Ablation: All stages – fidelity versus temporal stability.} Stage 2 improves fidelity over Stage 1 but slightly increases content drift, while Stage 3 provides the best overall trade-off between reconstruction quality and temporal stability.}
    \begin{tabular}{lccc}
    \toprule
         Stage & PSNR $\uparrow$ & SSIM $\uparrow$ & $E^*$ $\downarrow$  \\
    \midrule
         Alignment (S1) &  20.038 &  0.759 &  9.088  \\
         Perceptual (S2) & 24.114 & 0.846 & 8.508 \\
         Fidelity (S3) & \textbf{27.630}   & \textbf{0.869} & \textbf{8.005} \\
    \bottomrule
    \end{tabular}
    \label{tab:s1vs2vs3_ablation}
    \vspace{-0.15in}
\end{table}

\begin{figure}[htbp]
    \centering
    \includegraphics[width=0.99\linewidth]{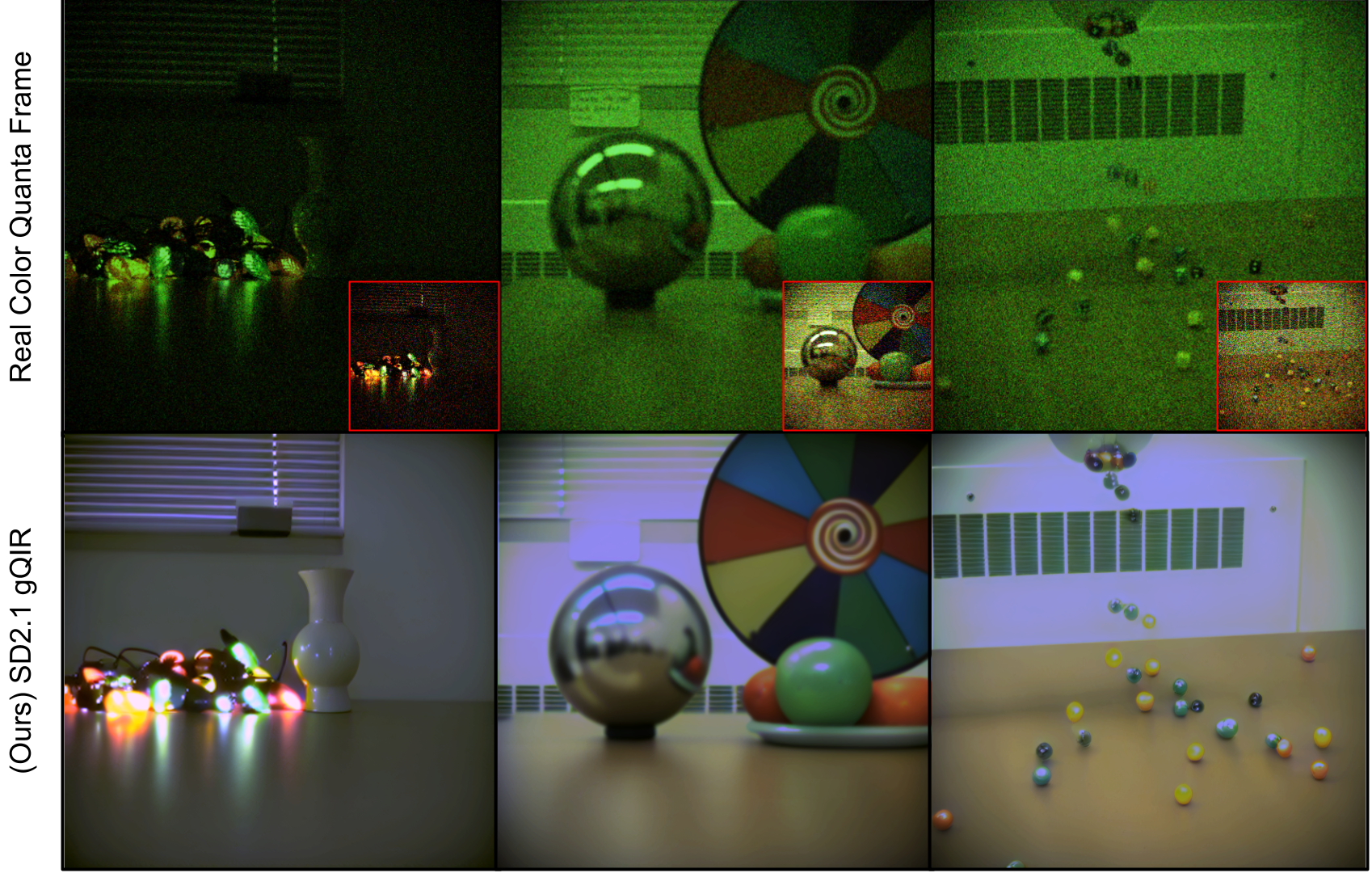}
    \caption{\textbf{Real color SPAD reconstructions.} Qualitative results on binary bursts captured with a 1Mpx passive color SPAD prototype at 6k fps. 
    Insets show demosaicing via sum-and-average.}
    \label{fig:real_captures}
    \vspace{-0.1in}
\end{figure}

\subsection{Real World Testing.}
gQIR reconstructs photorealistic images from real color SPAD captures without explicit correction for dark count or hot pixel as shown in~\cref{fig:real_captures}; the only post-processing applied is gray-world white balancing.
Interestingly, gQIR retains fidelity to the vignetting artifact inherent to our sensor prototype.
More qualitative results on our real-world acquisitions are provided in the supplementary material.

\vspace{-1mm}
\section{Limitations and Outlook}

This work presents the first use of large-scale generative priors for quanta burst reconstruction, introducing techniques tailored to emerging color SPAD sensors.
Despite resolution scalability via VAE tiling, several limitations remain. Motion cues from Stage 2 can degrade under subtle inter-frame drift, suggesting that video-level or multi-frame diffusion priors may further improve temporal coherence.
Second, our training assumes a fixed $3.5$~PPP, which limits robustness under extremely low-light (PPP $\le 1$). 
Explicitly modeling PPP as a conditioning signal may enhance generalization across lighting and sensor characteristics. 
Third, the 8-bit limit of the pretrained VAE decoder restricts the native HDR of SPADs~\cite{Ingle:CVPR21,Ingle:CVPR2019}; developing HDR-capable decoders is an important next step.



{
    \small
    \bibliographystyle{ieeenat_fullname}
    \bibliography{main}
}

\end{document}